\documentclass[runningheads]{llncs}
\usepackage{tablefootnote}
\usepackage{hyperref}
\usepackage[ruled]{algorithm2e}
\usepackage{color,xcolor}
\usepackage{bm}
\usepackage{amsmath}
\usepackage{tabularx}
\usepackage[percent]{overpic}
\usepackage{multirow}
\usepackage{cite}

\newcommand{\ignore}[1]{}

\begin{document}
%
\title{Improving 3D U-Net for Brain Tumor Segmentation by Utilizing Lesion Prior}
\titlerunning{Improving 3D U-Net for Brain Tumor Segmentation with Lesion Prior}
%
%

\author{Po-Yu Kao\inst{1}\and
Jefferson W. Chen\inst{2} \and
B.S. Manjunath\inst{1}}
\authorrunning{P.-Y. Kao et al.}
\institute{University of California, Santa Barbara, California, United States \\ \email{\{poyu\_kao, manj\}@ucsb.edu} \and
University of California, Irvine, California, United States
}


%
\maketitle
\setcounter{footnote}{0}
\begin{abstract}
We propose a novel, simple and effective method to integrate lesion prior and a 3D U-Net for improving brain tumor segmentation. First, we utilize the ground-truth brain tumor lesions from a group of patients to generate the heatmaps of different types of lesions. These heatmaps are used to create the volume-of-interest (VOI) map which contains prior information about brain tumor lesions. The VOI map is then integrated with the multimodal MR images and input to a 3D U-Net for segmentation. The proposed method is evaluated on a public benchmark dataset, and the experimental results show that the proposed feature fusion method achieves an improvement over the baseline methods. In addition, our proposed method also achieves competitive performance compared to state-of-the-art methods. 

\end{abstract}
\section{Introduction}
Primary central nervous system (CNS) tumors refer to a heterogeneous group of tumors arising from cells within the CNS and can be benign or malignant.
Malignant primary brain tumors remain among the most difficult cancers to treat, with a 5-year overall survival rate no greater than 35\%. 
The most common malignant primary brain tumors in adults are gliomas. 
In a patient with a suspected brain tumor, magnetic resonance imaging (MRI) with gadolinium is the investigation tool of choice \cite{lapointe2018primary}. 
Manual segmentation of brain tumors on MR images is a challenging and time-consuming task. 
Therefore, an automatic and accurate brain tumor segmentation tool benefits radiologists and physician on both diagnosis and treatment planning.

Convolutional neural networks have achieved state-of-the-art performance in the recent Multimodal Brain Tumor Image Segmentation Benchmarks (BraTS) \cite{isensee2017brain,isensee2018no,kamnitsas2017ensembles,myronenko20183d}.
These works focus on designing a new network architecture, loss function, data augmentation, and training and testing procedure in order to improve the performance of brain tumor segmentation.
Another method proposed by Kao et al. \cite{kao2018brain,10.3389/fnins.2019.01449} utilizes an existing brain parcellation to bring location information of brain into patch-based neural networks that improves the brain tumor segmentation performance of networks. 
Inspired by their work, we directly integrate lesion prior with multimodal MR images and input the fused information to a 3D U-Net. 
The proposed lesion prior fusion method includes two steps: (i) we first create a volume-of-interest (VOI) map from the ground-truth brain tumor lesions, and (ii) this VOI map is then integrated with the multimodal MR images and input to a 3D U-Net for the brain tumor segmentation.
The main contribution of this paper is the integration of lesion prior to a 3D U-Net architecture that improves the brain tumor segmentation performance of the 3D U-Net.
\section{Materials and Methods}
\subsection{Dataset}
Multimodal Brain Tumor Image Segmentation Benchmark (BraTS) 2017 \cite{bakas2017advancing,bakas2017gbm,bakas2017lgg,menze2015multimodal} provides 285 subjects in the training set and 46 subjects in the validation set. Multimodal MR images are provided for each subject, but ground-truth lesion mask is only available for the training subject. These MR images include T1-weighted, contrast-enhanced T1-weighted, T2-weighted, and fluid-attenuated inversion recovery scans, and the ground-truth lesion mask comprises the enhancing tumor (ET), edema (ED), and necrotic \& non-enhancing tumor (NCR/NET). The dimension of each image is $240\times240\times155$ in the $x, y$ and $z$ direction, and the voxel resolution is $1mm^{3}$. The provided data are intra-subject registered, interpolated to the same resolution and skull-stripped.
\subsection{Volume-of-interest Map} \label{set:VOI}

\begin{figure*}[htbp!]
    \centering
    \includegraphics[width=\textwidth]{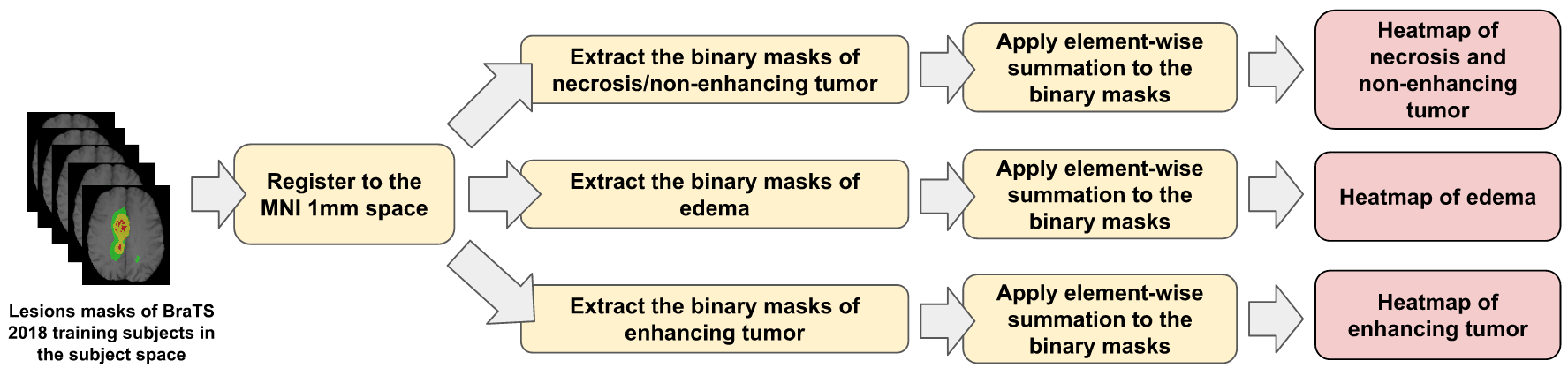}
    \caption{The workflow of building the heatmaps of different types of brain lesions.}
    \label{fig:heatmapsworkflow}
\end{figure*}

\begin{figure*}[htbp!]
    \centering
    \begin{overpic}[width=21mm, height=23.52mm]{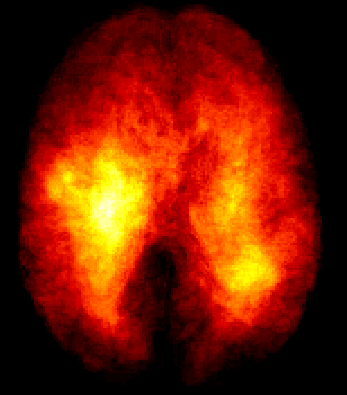}\color{white}\put(1,3){\textbf{ED}}\end{overpic}
	\begin{overpic}[width=21mm, height=23.52mm]{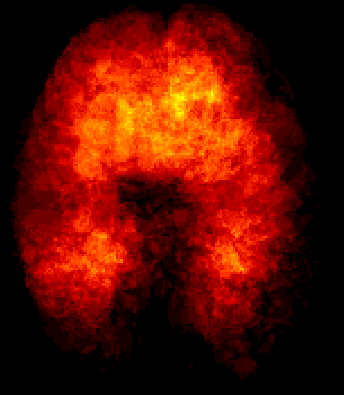}\color{white}\put(1,3){\textbf{NCR/NET}}\end{overpic}
	\begin{overpic}[width=21mm, height=23.52mm]{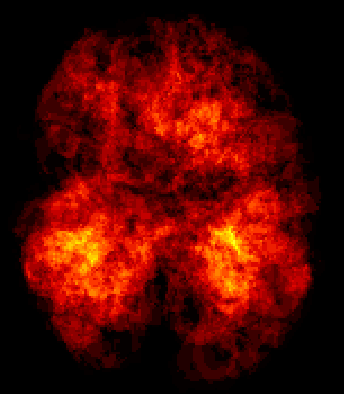}\color{white}\put(1,3){\textbf{ET}}\end{overpic}
    \caption{The heatmaps of different brain tumor lesions. The brighter voxels (yellow) represent higher intensity values. \textbf{Best viewed in color.}}
    \label{fig:heatmaps}
\end{figure*}

The volume-of-interest (VOI) map is built in the Montreal Neurological Institute (MNI) 1mm space \cite{grabner2006symmetric}, and each voxel of the VOI map has a label ranging from 0 to 9, which represents different probabilities of observing the brain tumor lesions.
First, we build the heatmaps of different types of brain tumor lesions in the MNI space, see the workflow in Fig.~\ref{fig:heatmapsworkflow}.
We apply inter-subject registration which registers the ground-truth lesions of each BraTS 2017 training subject from the subject space to the MNI space using FLIRT \cite{jenkinson2001global} from FSL. We then split the brain lesions of each subject into three binary masks, and each binary mask only contains information of one type of lesion. For each type of lesion, we apply element-wise summation to the binary masks of all 285 training subjects and create the heatmap of this type of lesion. Fig.~\ref{fig:heatmaps} shows the heatmaps of different brain tumor lesions from BraTS 2017 training subjects in the MNI space. 

The heatmaps of different brain lesions are then used to create the VOI map. 
The VOI map construction accounts for the fact that the whole tumor is a superset of ET, NCR/NET and ED, and the tumor core includes ET and NCR/NET. In addition, ETs are usually observed in patients with high-grade gliomas whose survival rate is considerably lower than patients with low-grade gliomas. Based on these observations, we create Algorithm~\ref{alg:voi} to generate the VOI map and prioritize the order of the VOI labels.


\begin{algorithm}[htbp!]
 \SetKwInOut{Input}{input}\SetKwInOut{Output}{output}
 \Input{A heatmap $H_{ed}$ of ED of size $w\times l \times d$ \\
 A heatmap $H_{ncr}$ of NCR/NET of size $w\times l \times d$ \\ 
 A heatmap $H_{et}$ of ET of size $w\times l \times d$} 
 \Output{The VOI map $V$ of size $w\times l \times d$ }
 $h_{ed,1}, h_{ed,2}, h_{ed,3} \leftarrow  \alpha, \beta, \gamma$ percentile of non-zero voxels of $H_{ed}$\;
 $h_{ncr,1}, h_{ncr,2}, h_{ncr,3} \leftarrow \alpha, \beta, \gamma$ percentile of non-zero voxels of $H_{ncr}$\;
 $h_{et,1}, h_{et,2}, h_{et,3} \leftarrow \alpha, \beta, \gamma$ percentile of non-zero voxels of $H_{et}$\;
 \For{$i\leftarrow 1$ \KwTo $w$}{
  \For{$j\leftarrow 1$ \KwTo $l$}{
   \For{$k\leftarrow 1$ \KwTo $d$}{
    \uIf{$H_{et}[i,j,k] \geq h_{et,3} $}{$V[i,j,k] \leftarrow 9$\;}
    \uElseIf{$H_{ncr}[i,j,k] \geq h_{ncr,3} $}{$V[i,j,k] \leftarrow 8$\;}
    \uElseIf{$H_{ed}[i,j,k] \geq h_{ed,3} $}{$V[i,j,k] \leftarrow 7$\;}
    \uElseIf{$H_{et}[i,j,k] \geq h_{et,2} $}{$V[i,j,k] \leftarrow 6$\;}
    \uElseIf{$H_{ncr}[i,j,k] \geq h_{ncr,2} $}{$V[i,j,k] \leftarrow 5$\;}
    \uElseIf{$H_{ed}[i,j,k] \geq h_{ed,2} $}{$V[i,j,k] \leftarrow 4$\;}
    \uElseIf{$H_{et}[i,j,k] \geq h_{et,1} $}{$V[i,j,k] \leftarrow 3$\;}
    \uElseIf{$H_{ncr}[i,j,k] \geq h_{ncr,1} $}{$V[i,j,k] \leftarrow 2$\;}
    \uElseIf{$H_{ed}[i,j,k] \geq h_{ed,1} $}{$V[i,j,k] \leftarrow 1$\;}
    \Else{$V[i,j,k] \leftarrow 0$\;}
    }
   }
 }
 \caption{Build the VOI map from the heatmaps of lesions.}
 \label{alg:voi}
\end{algorithm}

\begin{figure*}[htpb!]
    \centering
    \begin{overpic}[height=29mm]{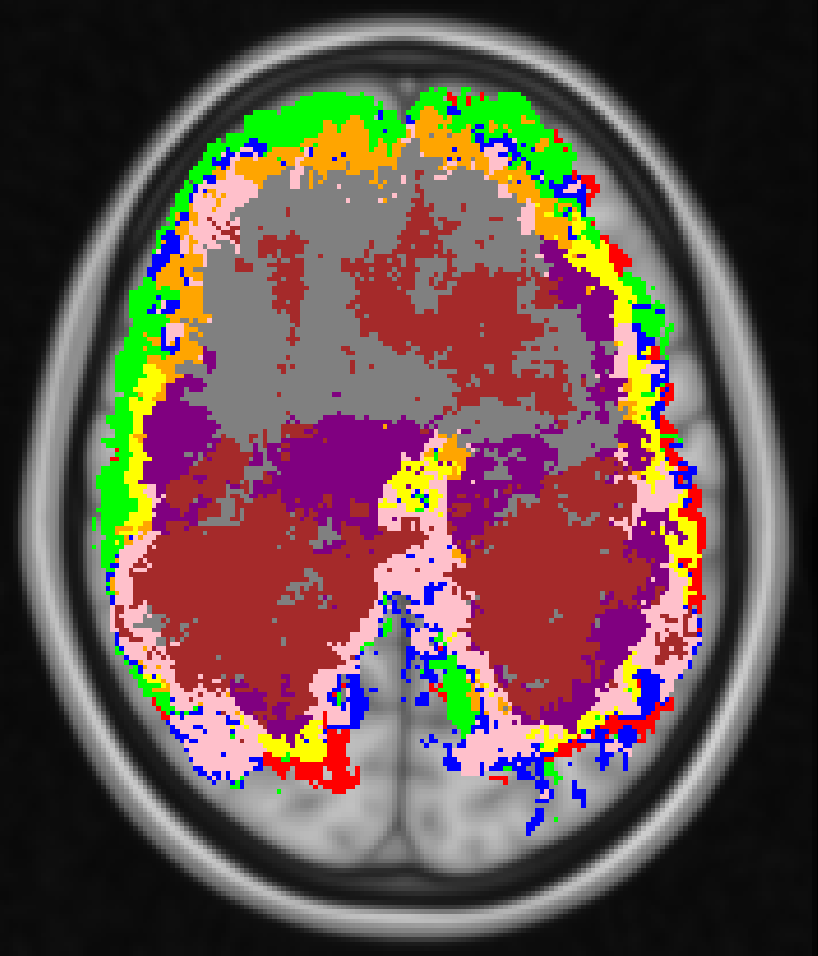}\color{white}\put(1,3){\textbf{VOI map}}\end{overpic}
	\begin{overpic}[height=29mm]{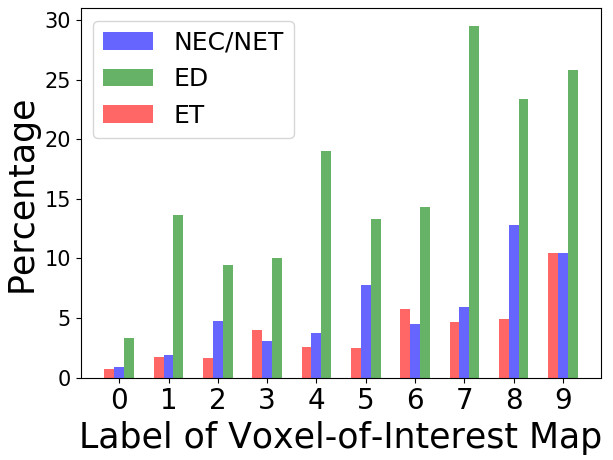}\end{overpic}
	\caption{The VOI map (background-0, red-1, green-2, blue-3, yellow-4, orange-5, pink-6, purple-7, grey-8, and brown-9)  and the distribution of brain tumor lesions (green-ED, blue-NEC/NET, and red-ET) observed in the different labels of VOI map from BraTS 2017 training subjects. \textbf{Best viewed in color.}}
    \label{fig:voi}
\end{figure*}

Note that the VOI labels are based on the thresholds which are chosen from the percentiles of non-zero voxels of heatmaps. 
For each lesion type, we sort the frequency counts of the non-zero voxels, and the heatmaps are used to generate these frequency counts.
The percentile thresholds ($h_{ed}, h_{ncr}, h_{et}$) are selected from these sorted frequency counts. 
We then use these percentile thresholds to create the VOI label mapping.
Any given voxel location in the VOI map has probabilities of being different types of lesion. 
We examined different thresholds, and $(\alpha, \beta, \gamma) = (50, 65, 80)$ percentiles yield the best overall segmentation performance. Fig.~\ref{fig:voi} shows the VOI map and the distribution of brain tumor lesions occurring in the different labels of VOI map. This distribution is computed by dividing the total voxel value of lesions in the heatmaps by the total volume of the corresponding VOI label. This distribution shows that (i) the prior probabilities of different lesions depend on their corresponding labels in the VOI label map, and (ii) lesions have higher probabilities to happen in the larger VOI labels.

\subsection{3D U-Net}  \label{sec:unet}

\subsubsection{Data pre-processing.}
Intensity normalization is the procedure of mapping intensities of different MR images into a standard scale, and it is an essential step to avoid initial biases and improve the performance of the network. For each MR image, we first clip it at [0.2 percentile, 99.8 percentile] of non-zero voxels to remove the outliers and subsequently normalize every voxel within the brain with respect to their mean and standard deviation. That is, $\bar{x}_{i} = (x_{i}-\mu)/\sigma$ where $i$ is the index of voxel inside the brain, $\bar{x}_{i}$ is the normalized voxel, $x_{i}$ is the corresponding raw voxel, and $\mu$ and $\sigma$ are the mean and standard deviation of the raw voxels inside the brain, respectively.

\subsubsection{Network architecture.}
\begin{figure}
    \centering
    \includegraphics[width=\textwidth]{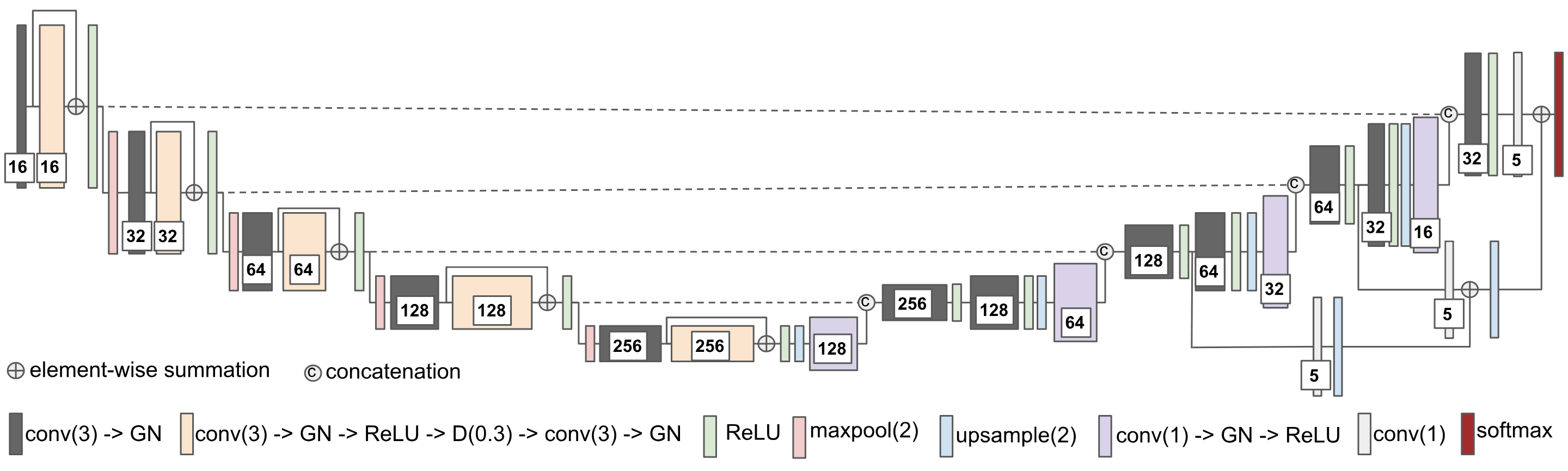}
    \caption{The proposed network architecture. conv(3): $3\times3\times3$ convolutional layer, GN: group normalization, D(0.3): dropout layer with 0.3 dropout rate, maxpool(2): $2\times2\times2$ max pooling layer, and conv(1): $1\times1\times1$ convolutional layer. \textbf{Best viewed in color.}}
    \label{fig:3dunet}
\end{figure}

The proposed network architecture shown in Fig.~\ref{fig:3dunet} is based on 3D U-Nets \cite{cciccek20163d,isensee2017brain}.
Different colors of blocks represent different types of layers. 
The number of convolutional kernels is indicated within the white box.
Group normalization \cite{wu2018group} is used, and the number of groups is set to 4.
Trilinear interpolation is used in the upsampling layer. 

\subsubsection{Training and testing procedure.}
The proposed network is trained with randomly cropped patches of size $128 \times 128 \times 128$ voxels and batch size 2. 
A larger input patch capture more contextual information of the brain. 
In every epoch, a cropped patch is randomly extracted from each subject.
The network is trained for a total of 300 epochs. 
The weights of network are updated by Adam algorithm \cite{kingma2014adam} with an initial learning rate $l_{0} = 10^{-3}$ following the schedule of $l_{0} \times 0.1^{\text{epoch}}$, L2 penalty weight decay of $10^{-4}$, and AMSGrad \cite{reddi2018convergence}. 
For the loss function, the standard multi-class cross-entropy loss with the hard negative mining is used to solve the class imbalance problem of the dataset.
We only back-propagate the negative (background) voxels with the largest losses (hard negative) and the positive (lesions) voxels to the gradients. 
In our implementation, the number of selected negative voxels is at most three times more than the number of positive voxels.
In addition, data augmentation is not used for both training and testing. 
At the testing time, we input the entire image of size $240\times240\times155$ voxels into the trained 3D U-Net for each patient to get the predicted lesion mask.
Training takes approximate 12.5 hours, and testing takes approximate 1.5 seconds per subject on an Nvidia 1080 Ti GPU.

\subsection{Integrate the VOI Map and a 3D U-Net}

\begin{figure}
    \centering
    \includegraphics[width=0.75\textwidth]{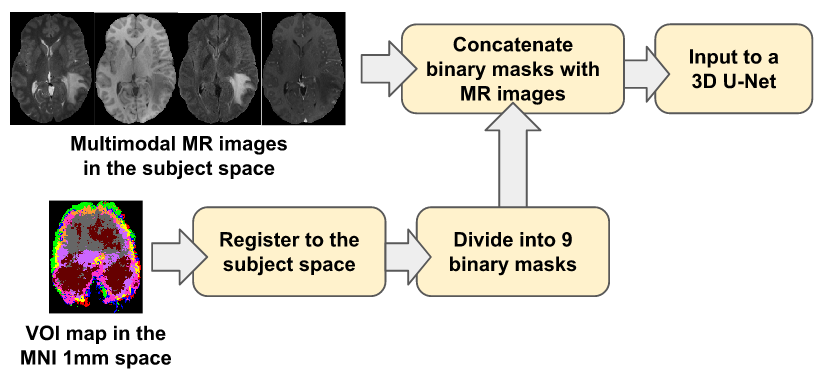}
    \caption{The pipeline of integrating the VOI map and a 3D U-Net.}
    \label{fig:integrate}
\end{figure}

Fig.~\ref{fig:integrate} shows the pipeline of integrating the VOI map and a 3D U-Net for brain tumor segmentation. 
First, we register the VOI map from the MNI 1mm space to the subject space using FLIRT \cite{jenkinson2001global} from FSL, and this registered VOI map is then split into 9 binary masks.
Each binary mask only contains information of one VOI label. 
Afterward, these binary masks are concatenated with the multimodal MR images. 
In the end, we input this 13-channel (4 image channels + 9 VOI channels) image to a 3D U-Net for both training and testing.

\subsection{Evaluation Metrics}
The employed evaluation metrics are the (i) Dice similarity coefficient (DSC) and the (ii) 95 percentile of the Hausdorff distance (H95).
DSC is the quotient of similarity and ranges between 0 and 1 which is defined as

\begin{equation*}
    \text{DSC} = \frac{2|G \cap P|}{|G|+|P|}
\end{equation*}
where $|G|$ and $|P|$ are the number of voxels in the ground-truth label and predict label, respectively.
Hausdorff distance $d_{H}(X,Y)$ measures how far two subsets $\{X, Y\}$ of a metric space are from each other which is defined as 
\begin{equation*}
    d_{H}(X,Y) = \max\{\underset{x\in X}{\sup} \: \underset{y\in Y}{\inf} \: d(x,y), \underset{y\in Y}{\sup} \: \underset{x\in X}{\inf} \: d(x,y)\}
\end{equation*}
where $d$ is the Euclidean distance, $\sup$ is the supremum, and $\inf$ is the infimum. 
\section{Experimental Results and Discussion}
First, we examine if the proposed lesion prior fusion method improves the brain tumor segmentation performance of the proposed 3D U-Net. 
Therefore, we train two identical 3D U-Nets with and without additional VOI map using 285 subjects of BraTS 2017 training set. 
BraTS 2017 validation set is used to evaluate the performance of these networks.
The quantitative results are shown in Table~\ref{tab:result}. 
From the first two rows of Table~\ref{tab:result}, our proposed lesion prior fusion method improves the performance of 3D U-Net, particularly for the DSC of ET (3.5\%), and H95 of ET (2.56) and whole tumor (2.39).

\begin{table}[htbp]
\caption{Quantitative results of the different models on BraTS 2017 validation set. Higher DSC and lower H95 indicate better segmentation performance. These results are given by the official online evaluation website. Results are reported as mean. Tumor core (TC) is the union of necrosis \& non-enhancing tumor and enhancing tumor (ET). Whole tumor (WT) is the union of edema, necrosis \& non-enhancing tumor and enhancing tumor. The underlined numbers highlight the improvement of VOI map, and the bold numbers highlight the best performance.}
\label{tab:result}
\centering
\begin{tabular}{ m{65mm}  c c c c c c}
& \multicolumn{3}{c}{DSC}  & \multicolumn{3}{c}{H95} \\ 
Model Descriptions & ET & WT  & TC & ET & WT & TC\\\hline
Single 3D U-Net (baseline) & 0.695 & 0.896 & 0.762 & 6.79 & 6.92 & 11.38 \\
Single 3D U-Net + VOI (proposed) & \underline{0.730} & \underline{0.899} & \underline{0.764} & \underline{4.23} & \underline{4.53} & \underline{10.93}\\\hline 
\rule{0pt}{2.5ex}Ensemble of five 3D U-Nets (baseline) & 0.723 & 0.902 & 0.763 & 5.99 & 4.75 & 10.58  \\
Ensemble of five 3D U-Nets + VOI (proposed)& \underline{\textbf{0.744}} & \underline{\textbf{0.903}} & \underline{0.780} & \underline{5.01} & \underline{\textbf{3.86}} & \underline{9.71}\\
Isensee et al. \cite{isensee2017brain} & 0.732 & 0.896 & \textbf{0.797} & 4.55 & 6.97 & 9.48 \\
Kamnitsas et al. \cite{kamnitsas2017ensembles} & 0.738 & 0.901 & \textbf{0.797} & \textbf{4.50} & 4.23 & \textbf{6.56} \\\hline
\end{tabular}
\end{table}

Second, we examine if the proposed lesion prior fusion method improves the performance of the ensemble of 3D U-Nets. 
Thus, we train two identical ensembles with and without additional VOI map using 285 subjects of BraTS 2017 training set. 
Each ensemble has five identical networks with different seed initializations, and the output of ensemble is averaged from five networks. 
BraTS 2017 validation set is used to evaluate the performance of ensembles, and the quantitative results are shown in Table~\ref{tab:result}.
From the middle two rows of Table~\ref{tab:result}, our proposed lesion prior fusion method also improves the tumor segmentation performance of the ensemble of five 3D U-Nets, particularly for the DSC of ET (2.1\%) and tumor core (1.7\%). 
The reason why the VOI map has the greatest improvement on the ET is that the percentiles of ET heatmap have the highest priorities while we create the VOI map.  
In addition, the proposed VOI map, directly built from the heatmaps of brain lesions, has inhomogeneous labels within neighboring voxels that carry more precise information of brain tumor lesions to the 3D U-Net. 

In the end, we compare the performance of our proposed method with the state-of-the-art methods \cite{isensee2017brain,kamnitsas2017ensembles}. 
From Table~\ref{tab:result}, the baseline model has worse performance than the state-of-the-art methods but it achieves a competitive performance by integrating the proposed VOI map. 
It is noted that the ensemble of Kamnitsas et al. \cite{kamnitsas2017ensembles} contains 7 different types of models but our proposed ensemble only consists of five 3D U-Net.

\section{Conclusion}
We have proposed a novel method to integrate prior information about the lesion probabilities into a 3D U-Net for improving brain tumor segmentation. Our experimental results demonstrate that the proposed lesion prior fusion approach improves the segmentation performance of the baseline model. 
Moreover, the proposed lesion prior fusion method can be easily integrated with other network architectures to further potentially enhance their segmentation performance.


\bibliographystyle{splncs04}
\bibliography{mybibfile}

%
\ignore{
\begin{algorithm}[htbp!]
 \SetKwInOut{Input}{input}\SetKwInOut{Output}{output}
 \Input{A heatmap $H_{ed}$ of ED of size $w\times l \times d$ \\
 A heatmap $H_{ncr}$ of NCR/NET of size $w\times l \times d$ \\ 
 A heatmap $H_{et}$ of ET of size $w\times l \times d$} 
 \Output{The VOI map $V$ of size $w\times l \times d$ }
 $h_{ed,1}, h_{ed,2}, h_{ed,3} \leftarrow $ 50, 65, 80 percentile of non-zero voxels of $H_{ed}$\;
 $h_{ncr,1}, h_{ncr,2}, h_{ncr,3} \leftarrow $ 50, 65, 80 percentile of non-zero voxels of $H_{ncr}$\;
 $h_{et,1}, h_{et,2}, h_{et,3} \leftarrow $ 50, 65, 80 percentile of non-zero voxels of $H_{et}$\;
 \For{$i\leftarrow 1$ \KwTo $w$}{
  \For{$j\leftarrow 1$ \KwTo $l$}{
   \For{$k\leftarrow 1$ \KwTo $d$}{
    \uIf{$H_{et}[i,j,k] \geq h_{et,3} $}{$V[i,j,k] \leftarrow 9$\;}
    \uElseIf{$H_{ncr}[i,j,k] \geq h_{ncr,3} $}{$V[i,j,k] \leftarrow 8$\;}
    \uElseIf{$H_{ed}[i,j,k] \geq h_{ed,3} $}{$V[i,j,k] \leftarrow 7$\;}
    \uElseIf{$H_{et}[i,j,k] \geq h_{et,2} $}{$V[i,j,k] \leftarrow 6$\;}
    \uElseIf{$H_{ncr}[i,j,k] \geq h_{ncr,2} $}{$V[i,j,k] \leftarrow 5$\;}
    \uElseIf{$H_{ed}[i,j,k] \geq h_{ed,2} $}{$V[i,j,k] \leftarrow 4$\;}
    \uElseIf{$H_{et}[i,j,k] \geq h_{et,1} $}{$V[i,j,k] \leftarrow 3$\;}
    \uElseIf{$H_{ncr}[i,j,k] \geq h_{ncr,1} $}{$V[i,j,k] \leftarrow 2$\;}
    \uElseIf{$H_{ed}[i,j,k] \geq h_{ed,1} $}{$V[i,j,k] \leftarrow 1$\;}
    \Else{$V[i,j,k] \leftarrow 0$\;}
    }
   }
 }
 \caption{Build the VOI map from the heatmaps of lesions.}
 \label{alg:voi}
\end{algorithm}
}

\end{document}